\documentclass{article}

\usepackage{microtype}
\usepackage{graphicx}
\usepackage{amsmath}
\usepackage{graphicx}
\usepackage{subcaption}
\usepackage{booktabs} 

\usepackage{hyperref}

\usepackage[accepted]{tccmlicml2021}

\icmltitlerunning{Deep Spatio-Temporal Forecasting of Electrical Vehicle Charging Demand}

\begin{document}

\twocolumn[
\icmltitle{Deep Spatio-Temporal Forecasting of Electrical Vehicle Charging Demand}

\icmlsetsymbol{equal}{*}

\begin{icmlauthorlist}
\icmlauthor{Frederik Boe Hüttel}{DTU}
\icmlauthor{Inon Peled}{DTU}
\icmlauthor{Filipe Rodrigues}{DTU}
\icmlauthor{Francisco C. Pereira}{DTU}
\end{icmlauthorlist}

\icmlaffiliation{DTU}{Department of Management, Technical University of Denmark, Lyngby, Denmark}

\icmlcorrespondingauthor{Fredeirk Boe Hüttel}{fbohu@dtu.dk}

\icmlkeywords{Machine Learning, ICML}

\vskip 0.3in
]



\printAffiliationsAndNotice{\icmlEqualContribution} 

\begin{abstract}
Electric vehicles can offer a low carbon emission solution to reverse rising emission trends. However, this requires that the energy used to meet the demand is green. To meet this requirement, accurate forecasting of the charging demand is vital. Short and long-term charging demand forecasting will allow for better optimisation of the power grid and future infrastructure expansions.  In this paper, we propose to use publicly available data to forecast the electric vehicle charging demand. To model the complex spatial-temporal correlations between charging stations, we argue that Temporal Graph Convolution Models are the most suitable to capture the correlations.  The proposed Temporal Graph Convolutional Networks provide the most accurate forecasts for short and long-term forecasting compared with other forecasting methods.
\end{abstract}

\section{Introduction}
Electric Vehicles (EVs) offer lower carbon emission when compared to gasoline-powered vehicles and can reverse rising emission trends assuming a green energy mix~\cite{Miotti2016, Coignard2018, Wei2021}.  As EV adoption increases and the charging network expands, forecasting the charging demand becomes increasingly essential for energy providers seeking to provide green energy to meet the demand~\cite{ML_Climate_change}. Operators of EV Charging Stations (EVCS) also have a stake in the forecast as they seek to find optimal locations for expansion of the infrastructure~\cite{en10101534,TAO2018735}. 

Forecasting of EV charging demand is generally well studied and explored in the literature, with methods ranging from queuing theory to machine and statistical learning~\cite{KHOO2014263, UKcase, en14051487, HAL_review}. Despite this interest in modelling the demand, the data used is often not publicly available. The lack of accessibility obscures results and reduces comparability. Fortunately, we find that comments on \emph{lack} of real-world public data are inaccurate~\cite{en14051487}. As highlighted by \cite{HAL_review}, there exist multiple open datasets, which are still underutilized in research. 

These public datasets provide sufficient data for research and forecasting. The datasets contains both diverse spatial and temporal resolutions. Prominent examples include the city of Palo Alto, U.S. which is on a dense spatial resolution with observations going back to 2011~\cite{PaloAlto}, and the city of Perth, Scotland which spans a spatially sparse area with observations from 2016 to 2019~\cite{perth}. Two other prominent datasets are from Boulder, U.S. \cite{colorado} and Dundee, Scotland~\cite{cityofdundee_2019}, which again spans a different spatial and temporal resolutions. We propose to use the publicly available data from to forecast the EV charging demand.

One of the challenges in forecasting the EV charging demand is the complex spatial and temporal dependencies between EVCS~\cite{HAL_review}. We argue that forecasts of EV charging demand must account for the dependencies between the stations, however, this dependence is often neglected by modelling the entire system of EVCS jointly~\cite {en14051487}. In recent years there has been increased research into Graph Convolution Network (GCN) which can handle graph-structured data~\cite{GCN} and extensions that model temporal graph-structured data~\cite{TGCN}. We leverage these temporal graph convolutions to capture the complex spatio-temporal correlations of the EVCS. The use of temporal GCN is, to our knowledge, previously unexplored in the domain of EV charging demand forecasting.

\section{Related work}
We now provide a brief overview of related works. For a more in-depth review of current publicly available datasets and models, the reader is referred to \cite{HAL_review} and the \emph{related work} section in \cite{en14051487}. 

\paragraph{Temporal forecasting}
Related studies can be broken down as either classical statistical or machine learning forecasting methods. The classical statistical models often providing interpretable parameters such as regression~\cite{en10101534} and ARIMA models~\cite{AMINI2016378}. Machine learning methods, such as random forest~\cite{MAJIDPOUR2016134, en11113207} and deep learning models~\cite{8476758, en14051487}, can provide higher forecast accuracy, at the expense of interpretability.

\paragraph{Spatio-temporal forecasting} 
In recent years there has been an increase in spatio-temporal forecasting with multiple methods for extraction of spatial and temporal features. Research has focused on raster maps above areas and using convolutional neural networks (CNN)~\cite{LeCun2015} to extract spatial features and recurrent deep learning layers to learn temporal features~\cite{rodrigues2018expectation}. The convolution operators have been extended to work on non-euclidean data using using graph convolutions and graph convolutional networks (GCN)~\cite{GCN}. The GCN models have seen some use in forecasting the demand of EVCS in a shared fleet of EVs~\cite{Luo2019DynamicDP}. Recent research have extendedn the GCN to model temporal correlations by incorporation of recurrent deep learning layers into the GCN (T-GCN)~\cite{TGCN}.

\section{Methodology}
\label{sec:method}
Incorporating the spatio-temporal correlations into the forecast should provide better estimates of the EV charging demand compared to methods that only consider a single charging station or all the charging stations jointly as one time-series. 

\paragraph{Problem definition} 
Given the temporal signal $X=\{\mathbf{x}_1,\mathbf{x}_2,\cdots,\mathbf{x}_t\}$ containing the historical daily energy demand for each charging station in an area and a topology $G$, which models the spatial correlations of the charging stations, the problem of spatio-temporal forecasting can be considered as learning a a function $f$ on topology $G$ with temporal signal $X$.
\begin{equation}
    \left[X_{t+1},\cdots,X_{t+T}\right] = f(G;(X_{t-n},X_{t-1},X_{t}))\,,
\end{equation}
where $n$ is the length of the historical signal and $T$ is the forecast horizon. $G$ can be presented through a \emph{Raster map} or with a \emph{Graph}, as follows.

\paragraph{Raster-map} In this case, the topology $G$ is a grid spanning the entire area. Each grid cell then pertains to a specific spatial location in the area, and corresponds to the sum of daily energy demand for stations in the cell. The forecast using a raster map is the daily demand in each grid cell.

\paragraph{Graph} In this cases, the topology is a graph $G=(V,E)$, where $V=\{v_1,v_2,..,v_N\}$ is the set of charging stations in the area. $N$ is the number of nodes and $E$ is the set of edges between the nodes in $V$. The forecast is then for the daily energy demand for each individual charging station. To model the correlation between individual charging stations, the weight of an edge $e_{i,j}$ is computed as:
\begin{equation}
e_{ij}=\begin{cases}
			\exp(-h(x_i,x_j))& \text{if $h(x_i,x_j) < 2.5$}\\
            0 & \text{otherwise}
		 \end{cases}\,,
\end{equation}
where $x_i$ and $x_j$ are the location of the charging stations and $h$ is the Haversine Distance in km. We can setup the adjacency matrix $A \in R^{N\times N}$ using $E$~\cite{TGCN}. Stations far away from each-other is assumed to no be very correlated, which makes us set edges for large distance to be non-existent.

\begin{figure}[tb]
\centering
\begin{subfigure}{.5\columnwidth}
  \centering
  \includegraphics[width=.9\columnwidth]{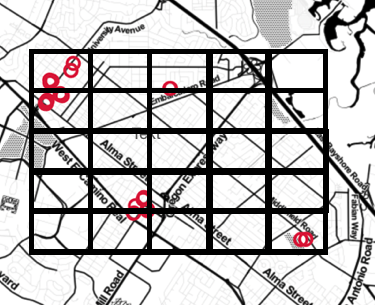}
  \caption{Raster map}
  \label{fig:sub1}
\end{subfigure}%
\begin{subfigure}{.5\columnwidth}
  \centering
  \includegraphics[width=.9\columnwidth]{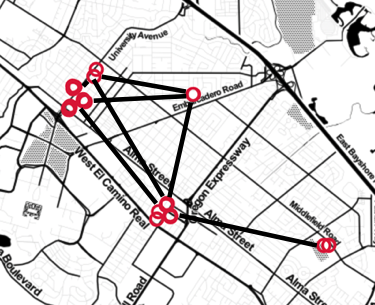}
  \caption{Graph}
  \label{fig:sub2}
\end{subfigure}
\caption{Figures showing the 2 topologies. Charging stations locations are marked with red circles. \autoref{fig:sub1} A raster map over the city of Palo Alto. \autoref{fig:sub2} Graph over charging stations in the city of Palo Alto, excluding edges within every station to highlight the overall graph structure.}
\label{fig:test}
\end{figure}

\subsection{Modelling}
For modelling the spatial and temporal features with the raster maps, a CNN can obtain spatial and temporal correlation by stacking historical raster maps in the channel dimension, making each grid cell a time-series of demand for the spatial location.

\paragraph{Long short term memory}
As an alternative to stacking raster maps, spatial features can be extracted from each raster map using a CNN. Then each grid cell can be stacked as a time-series and propagated through a long short term memory cell (CNN+LSTM)~\cite{LSTM}. The CNN and LSTM then capture the spatial and temporal features, respectively.

\paragraph{Temporal Graph Convolutional Network}
The spatial features the CNN captures will be heavily influenced by the size of the grid cells. If the grid is sufficiently fine-grained, then the CNN captures the topological structure, but at the expense of many grid cells with zero observations. On the other hand, if the grid cells are made too large, the model will not capture the complex topological structure. Conversely, a Graph Convolutional Network (GCN) can capture the complex topological structure with graph convolutions. The GCN constructs a filter in the Fourier domain, which acts on each graph node and its first-order neighbourhood to capture spatial dependencies between nodes~\cite{GCN}. A 2-layered GCN model is formulated as:
\begin{equation}
\label{eq:GCN}
 f(A,X)=\sigma\left(\widehat{A}\, \operatorname{Relu}\left(\widehat{A} X W_{0}\right) W_{1}\right)\,,
\end{equation}
where $X$ is the feature matrix, $A$ is the adjacency matrix, $\widehat{A}=\widetilde{D}^{-\frac{1}{2}} \widetilde{A} \widetilde{D}^{-\frac{1}{2}}$ denotes a preprocessing step, $\widehat{A}=A+I_{N}$ is a matrix with self-connection structure, and $\widetilde{D}$ is a degree matrix, $\widetilde{D}=\sum_{j}\widetilde{A}_{i j}$ . $W_0$ and $W_1$ are the weight matrices in the first and second layers, respectively, and $\sigma(\cdot)$, $\operatorname{Relu}()$ is the activation function.

The GCN extends to account for a temporal signal by combining the GCN with LSTM layers (T-GCN)~\cite{TGCN}. The key equations of the T-GCN with a LSTM cell can be summarised as follows, where $f(A,X_t)$ is the graph convolution from \autoref{eq:GCN}:
\begin{equation}
i_{t} =\sigma_{g}\left(W_{i} f(A,X_t)+U_{i} h_{t-1}+b_{i}\right)\,,
\end{equation}
\begin{equation}
f_{t} =\sigma_{g}\left(W_{f} f(A,X_t)+U_{f} h_{t-1}+b_{f}\right)\,,
\end{equation}
\begin{equation}
o_{t} =\sigma_{g}\left(W_{o} f(A,X_t)+U_{o} h_{t-1}+b_{o}\right)\,,
\end{equation}
\begin{equation}
\tilde{c}_{t} =\sigma_{c}\left(W_{c} f(A,X_t)+U_{c} h_{t-1}+b_{c}\right)\,,
\end{equation}
\begin{equation}
c_{t} =f_{t} \odot c_{t-1}+i_{t} \odot \tilde{c}_{t}\,,
\end{equation}
\begin{equation}
h_{t} =o_{t} \odot \sigma_{h}\left(c_{t}\right)\,.
\end{equation}
The matrices $W_{i-c}$ and $U_{i-c}$ contain the trainable weights, $\odot$ denotes the Hadamard product and $c_0=0$ and $h_0=0$. In summary, the T-GCN models capture both the complex topological structure of the charging stations, using the GCN, and the temporal structure of the data, using the LSTM layers.

\subsection{Loss function}
We denote the forecasted value as $\widehat{X}_{t+1:t+T}$ and the realised value as $X_{t+1:t+T}$, and aim to minimise forecast errors. The models are optimized using the \emph{mean absolute error} loss and stochastic gradient descent. The networks are regularised through $\ell_2$  regularisation of the parameter, making the loss:
\begin{equation}
\label{eq:loss}
    \mathcal{L}=|X_{t+1:t+T}-\widehat{X}_{t+1:t+T}|+ \lambda \beta^2
    \,,
\end{equation}
where $\lambda$ is a tuneable hyperparameter and $\beta$ denotes the parameters of the models.
\section{Data}
\begin{figure}[tb]
\vskip 0.2in
\begin{center}
\centerline{\includegraphics[width=\columnwidth]{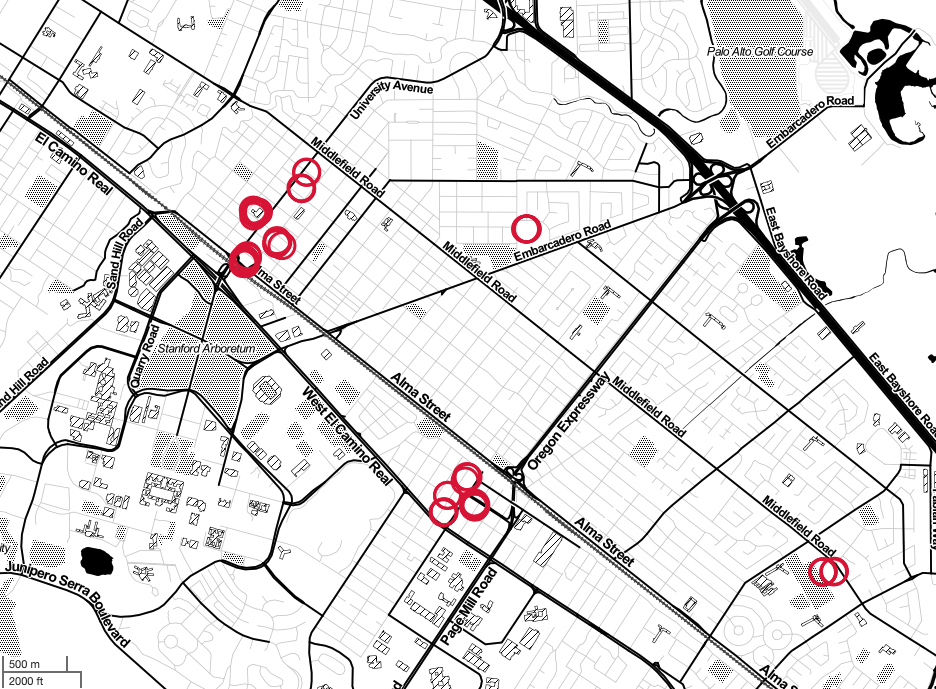}}
\caption{Palo alto and the location of EV chargers}
\label{fig:PaloAlto}
\end{center}
\vskip -0.4in
\end{figure}

The proposed models are empirically evaluated using data from the city of Palo Alto \cite{PaloAlto}. The data consists of EV charging transactions at the locations depicted in \autoref{fig:PaloAlto}. The data set contains various meta-data on the charging transaction such as Gasoline Savings, Charging time, Plug type. The meta-data is excluded from the analysis, because it is not available for the other public datasets. We focus on using purely the energy consumption (kWh) for a transaction since this is computeable for all datasets, making the models applicable to those. The consumption is aggregated into a a daily energy demand for each of the stations in Palo Alto. For the raster map, we use a grid of $5 \times 5$ and for the graph is created as explained in \autoref{sec:method}. The data from Palo Alto is continually updated and contains data up until 2021.
\section{Experiments}
The models are tested on three different forecasts horizons; 1 day, 7 day and 30 day. The different horizons thus reflect various usages of the models, as follows. The long forecast horizon of 30 days can be used by EVCS operators looking to forecast usage of their system. The short horizons can be used by energy providers to plan for and optimise the dayahead energy consumption, such that they can meet the demand with clean energy. Models for the 1 and 7 day forecasts are trained using the last 30 days. For the 30 days forecast, the models use the last 120 days. The models are implemented in \textit{keras}~\cite{keras} and the graphs are implemented using \textit{stellar graph}~\cite{StellarGraph}. The CNN uses 16 kernels to forecast based on the last 30 days raster maps. The CNN+LSTM uses 16 kernels in the CNN and 50 hidden units in the LSTM. The T-GCN uses 16 and 10 filters along with 50 hidden units in the LSTM. All models are fitted using the loss in  \autoref{eq:loss} with $\lambda=10^{-3}$ and the ADAM optimiser~\cite{Adam} for 1000 epochs\footnote{The code is public at \href{https://github.com/fbohu/Deep-Spatio-Temporal-Forecasting-of-Electrical-Vehicle-Charging-Demand}{Github}}. 

\begin{figure*}[tb]
\centering
\begin{subfigure}{.5\textwidth}
  \centering
  \includegraphics[width=.9\columnwidth]{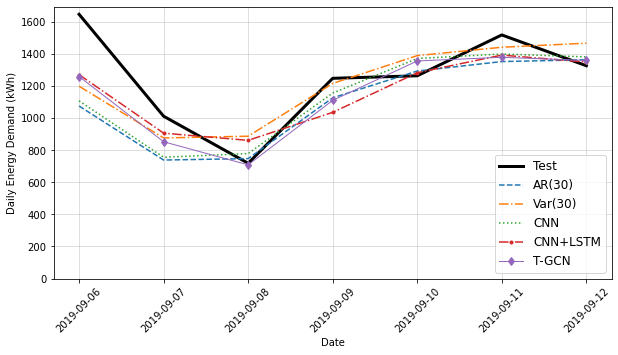}
  \caption{7 day forecast}
  \label{fig:7day}
\end{subfigure}%
\begin{subfigure}{.5\textwidth}
  \centering
  \includegraphics[width=.9\columnwidth]{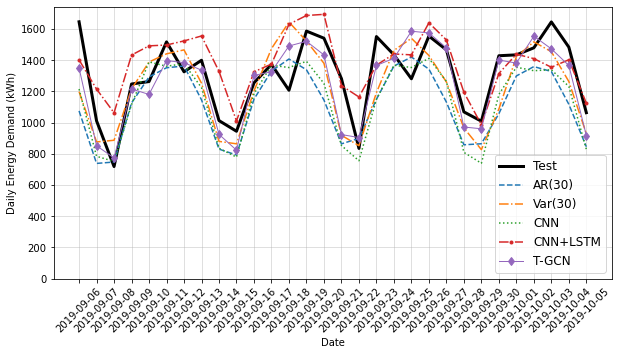}
  \caption{30 day forecast}
  \label{fig:30day}
\end{subfigure}
\caption{Visualisation of the forecast from the various models. Each model appears to have captured the temporal correlations of the system. In \autoref{fig:7day} and \autoref{fig:30day} we show forecasts from the different models with different horizons. From the 30 day forecast a weekly pattern from all the forecasts can be observed.}
\label{fig:forecasts}
\end{figure*}

\subsection{Evaluation}
We evaluate the models on the forecast daily energy demand for the \textit{total system}, which is the sum of all the energy needed for the area.  For the T-GCN model, it is the sum of the forecasts for each station. For the CNN and CNN+LSTM it is the sum for each grid cell. We compare the CNN, CNN+LSTM and T-GCN with a baseline Auto-regressive (AR) model fitted on the entire system and a baseline of a vector auto-regressive (VAR). The VAR model uses the grid cells from the raster map as inputs. Both baselines models use a model order of 30, in order to match the 30 lagged values as the deep learning model are trained with. We train the models with data between 2012 and 2019.

\subsection{Experimental Results}
We evaluate the models with the \textit{Root mean squared error} (RMSE) across the different forecast horizons (\autoref{tab:forecast_results}). As can be expected, finer spatial resolution yields higher predictive quality. Since the LSTMs in both the T-GCN and the CNN+LSTM are the same, the graph convolutions are superior to convolutions at extracting features required for the forecast.
\begin{table}[b]
\caption{Average RMSE from 3 runs. Uncertainty in the scores is  represented as the standard deviation over the 3 runs.}
\label{tab:forecast_results}
\vskip 0.15in
\begin{center}
\begin{small}
\begin{sc}
\begin{tabular}{lrrr}
\toprule
Model & 1 day & 7 days  & 30 days\\
\midrule
AR(30) &  178 & 251 & 252 \\
VAR(30) & 189 & 203 & 201   \\
CNN & $144 \pm 12$ & $243  \pm 4$ & $211 \pm 4$ \\
CNN+ LSTM & $95\pm 11$ & $192\pm 8$ & $187 \pm 7$\\
T-GCN & $\mathbf{61 \pm 8}$ & $\mathbf{184 \pm 9}$ & $\mathbf{161 \pm 15}$ \\

\bottomrule
\end{tabular}
\end{sc}
\end{small}
\end{center}
\vskip -0.1in
\end{table}
Across the different forecast horizons, the T-GCN provides the best forecasting of the models. For both the short and long term predictions, the T-GCN is showing a lower RMSE than the other proposed methods. The graph is able to capture the small difference that is present in each EVCS, as opposed to the raster map presentation, and to capture the correlations between the EVCS. \autoref{fig:forecasts} shows the forecast for the different models. All models appear to capture the weekly patterns, where the demand is lower in the weekends, compared to weekdays.

\section{Conclusion}
 Throughout this paper, we have argued for the use of publicly available data for forecasting the electric vehicle charging demand. Based on the experimental results, Graph Convolutional Networks have superior forecasting performance compared to other methods. In particular, we see that the Graph Convolutional Network model can capture the spatial and temporal correlations of the network of charging stations better than plain convolutional neural networks.

The models presented in this work have assumed a known expansion of the charging station network. In future work, we will incorporate dynamically evolving graphs. Further, previous research in demand forecasting has shown that the observed demand is probably \emph{censored}, e.g., by the maximum capacity of the charging stations or through demand lost to competing charging services, which makes the true demand latent. There exist various ways to handle censored data which we intend to explore in future work~\cite{Biganzoli, Greene, GAMMELLI2020102775}.

We hope that the results and arguments encourage researchers to use publicly available data for research into EV charging demand. In particular, we encourage researchers and stakeholders in EV charging to use Graph Convolutional Networks to forecast the electric vehicle charging demand.  We believe the methods provided here can help expand the electric vehicle charging infrastructure and help reverse rising emissions trends.


\bibliography{refs}
\bibliographystyle{tccmlicml2021}


\end{document}